\documentclass[letterpaper, 10 pt, conference]{ieeeconf}

\IEEEoverridecommandlockouts
\usepackage{afterpage}
\usepackage{xspace}
\usepackage{amssymb,paralist,epsfig,standalone,bm,placeins}  
\usepackage{graphicx} 
\usepackage{xcolor}
\usepackage{multirow}
\usepackage{makecell}
\usepackage{multicol}
\usepackage[utf8]{inputenc}
\usepackage{amsmath,amsfonts,booktabs,cite} 
\usepackage{siunitx,textcomp} 
\usepackage[hidelinks]{hyperref} 
\let\labelindent\relax
\usepackage[inline]{enumitem} 
\usepackage[nolist]{acronym} 
\usepackage{caption}
\usepackage{tikz,standalone,pgfplots}
\usetikzlibrary{patterns}
\usepackage{pgfplots, pgfplotstable}
\graphicspath{{figures_tex/}}

\usepackage{changes}
\usepackage[ruled,vlined]{algorithm2e}

\makeatletter
\newcommand{\removelatexerror}{\let\@latex@error\@gobble}
\makeatother
\pgfplotsset{compat=1.12}
\newacro{mss}[MSS]{Mass-Spring System}
\newacro{pbd}[PBD]{Position-based Dynamics}
\newacro{fem}[FEM]{Finite Element Method}
\newacro{dnn}[DNN]{Deep Neural Network}
\newacro{fcn}[FCN]{fully-convolutional network}

\newcommand{\figref}[1]{\hyperref[#1]{Fig.~\ref*{#1}}}
\newcommand{\tabref}[1]{\hyperref[#1]{Table~\ref*{#1}}}
\newcommand{\secref}[1]{\hyperref[#1]{Section~\ref*{#1}}}
\newcommand{\algoref}[1]{\hyperref[#1]{Algorithm~\ref*{#1}}}

\newcommand{\ra}[1]{\renewcommand{\arraystretch}{#1}}
\newcommand{\tbs}[1]{\renewcommand{\tabcolsep}{#1pt}}

\def\bestcolor{(best viewed in color)}
\def\panda{Franka Emika Panda\xspace}

\def\etal{\textit{et al.} }

\def\figvspace{\vspace{-1.2em}}

\definecolor{findOptimalPartition}{HTML}{D7191C}
\definecolor{storeClusterComponent}{HTML}{FDAE61}
\definecolor{dbscan}{HTML}{ABDDA4}
\definecolor{constructCluster}{HTML}{2B83BA}

\title{\LARGE \bf
SPONGE: Sequence Planning with Deformable-ON-Rigid Contact Prediction from Geometric Features}

\author{Tran~Nguyen~Le$^{1}$, Fares~J.~Abu-Dakka$^{2}$, Ville~Kyrki$^{1}$%
\thanks{This work was supported by CHIST-ERA project IPALM (326304). Abu-Dakka is supported by the European project euROBIN under grant agreement No 101070596. We gratefully acknowledge the support of NVIDIA Corporation with the donation of the Titan Xp GPU used for this research.} \thanks{$^{1}$ Intelligent Robotics Group at the Department of Electrical Engineering and
Automation, School of Electrical Engineering, Aalto University, Finland.
\texttt{\{firstname.lastname\}{@}aalto.fi}}
\thanks{$^{2}$ MIRMI, Technische Universität München, 80992 München, Germany. Part of the research presented in this work was conducted when F.\ Abu-Dakka was at Aalto University. \texttt{fares.abu-dakka{@}tum.de}
}}

\begin{document}
\maketitle
\thispagestyle{empty}
\pagestyle{empty}


\begin{abstract}
Planning robotic manipulation tasks, especially those that involve interaction between deformable and rigid objects, is challenging due to the complexity in predicting such interactions. We introduce SPONGE, a sequence planning pipeline powered by a deep learning-based contact prediction model for contacts between deformable and rigid bodies under interactions. The contact prediction model is trained on synthetic data generated by a developed simulation environment to learn the mapping from point-cloud observation of a rigid target object and the pose of a deformable tool, to 3D representation of the contact points between the two bodies. We experimentally evaluated the proposed approach for a dish cleaning task both in simulation and on a real \panda with real-world objects. The experimental results demonstrate that in both scenarios the proposed planning pipeline is capable of generating high-quality trajectories that can accomplish the task by achieving more than 90\% area coverage on different objects of varying sizes and curvatures while minimizing travel distance. Code and video are 
available at: \url{https://irobotics.aalto.fi/sponge/}.
\end{abstract}

\section{Introduction}
\label{sec:introduction}
Volumetric or 3D deformable objects, present in various forms, are prevalent in many aspects of daily life, including food, toys, or internal organs of humans. Successful manipulation of such objects can lead to numerous practical applications in areas such as surgical manipulation, or food processing where robots can be used to make pizza dough \cite{lin2022planning}, cut fruits \cite{heiden2021disect}, or in healthcare where robots can be used to rearrange objects in target configurations \cite{shen2022acid}, or clean dishes \cite{wang2022visual}. These tasks are trivial for humans because not only we possess remarkable dexterity but we also excel at task planning, as demonstrated by our ability to perceive objects at hand, and develop a plan to complete the task with precision and accuracy in less than a second.


\begin{figure}[!t]
    \centering
    \def\svgwidth{\linewidth}
     {\fontsize{6}{6}
\begingroup%
  \makeatletter%
  \providecommand\color[2][]{%
    \errmessage{(Inkscape) Color is used for the text in Inkscape, but the package 'color.sty' is not loaded}%
    \renewcommand\color[2][]{}%
  }%
  \providecommand\transparent[1]{%
    \errmessage{(Inkscape) Transparency is used (non-zero) for the text in Inkscape, but the package 'transparent.sty' is not loaded}%
    \renewcommand\transparent[1]{}%
  }%
  \providecommand\rotatebox[2]{#2}%
  \newcommand*\fsize{\dimexpr\f@size pt\relax}%
  \newcommand*\lineheight[1]{\fontsize{\fsize}{#1\fsize}\selectfont}%
  \ifx\svgwidth\undefined%
    \setlength{\unitlength}{8842.60986328bp}%
    \ifx\svgscale\undefined%
      \relax%
    \else%
      \setlength{\unitlength}{\unitlength * \real{\svgscale}}%
    \fi%
  \else%
    \setlength{\unitlength}{\svgwidth}%
  \fi%
  \global\let\svgwidth\undefined%
  \global\let\svgscale\undefined%
  \makeatother%
  \begin{picture}(1,0.42238661)%
    \lineheight{1}%
    \setlength\tabcolsep{0pt}%
    \put(0,0){\includegraphics[width=\unitlength,page=1]{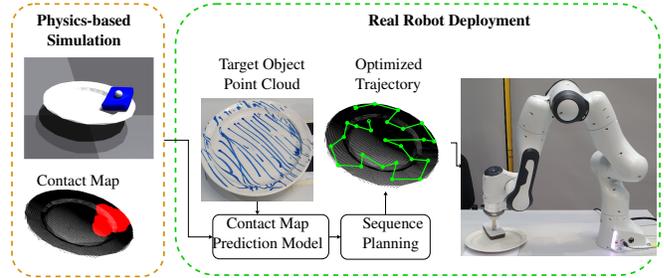}}%
    \put(0.54103089,0.06){\makebox(0,0)[lt]{\lineheight{1.25}\smash{\begin{tabular}[c]{l}Sequence \\ Planning\end{tabular}}}}%
    \put(0.31,0.075){\makebox(0,0)[lt]{\lineheight{1.25}\smash{\begin{tabular}[t]{c}Contact Map \\ Prediction Model\end{tabular}}}}%
    \put(0.32,0.32){\makebox(0,0)[lt]{\lineheight{1.25}\smash{\begin{tabular}[t]{c}Target Object \\ Point Cloud\end{tabular}}}}%
    \put(0.53,0.32){\makebox(0,0)[lt]{\lineheight{1.25}\smash{\begin{tabular}[t]{l}Optimized \\ Trajectory\end{tabular}}}}%
    \put(0.55,0.38717223){\makebox(0,0)[lt]{\lineheight{1.25}\smash{\begin{tabular}[t]{l}\textbf{Real Robot Deployment}\end{tabular}}}}%
    \put(0.04,0.38717223){\makebox(0,0)[lt]{\lineheight{1.25}\smash{\begin{tabular}[t]{c}\textbf{Physics-based} \\ \textbf{Simulation}\end{tabular}}}}%
    \put(0.04,0.14195381){\makebox(0,0)[lt]{\lineheight{1.25}\smash{\begin{tabular}[t]{l}Contact Map\end{tabular}}}}%
  \end{picture}%
\endgroup%
} 
    \caption{SPONGE deployed in the real world to accomplish a dish cleaning task with a deformable sponge. Given a point cloud of the target objects, SPONGE powered by a contact map prediction model trained in simulation, plans an optimal trajectory aiming at achieving full area coverage of target objects with the least number of waypoints.} 
    \label{fig:title}
    \vspace{-0.5cm}
\end{figure}

Planning manipulation tasks involving interactions between deformable and rigid objects, such as wiping a curved surface with a deformable tool, is difficult due to the challenge in predicting such interactions. To date, most existing works disregard the interaction between the deformable tool and target objects, and focus only on the control aspect of the tasks \cite{control_survey}. Only recently have some researchers started to investigate how to estimate and harness such interactions in different tasks such as assistive dressing, or food processing. In the literature, researchers studied the interaction between complex deformable objects such as human hands \cite{Brahmbhatt_2020_ECCV, jiang2021graspTTA} or cloth \cite{wang2022visual,erickson17} and rigid bodies by looking at the concept of \textit{contact reasoning} where the location of contact and the magnitude of applied forces are estimated once the two bodies interact with each other. However, this \textit{contact reasoning} concept is more suitable for the control aspect than for the planning aspect due to its ability to track the interaction in real time. Thus, the question of how to predict the interaction between deformable and rigid objects and exploit such interactions for planning remains open.

To address the aforementioned open issues, we propose \textbf{S}equence \textbf{P}lanning with deformable-\textbf{ON}-rigid contact prediction from \textbf{GE}ometric features (SPONGE), a sequence planning pipeline powered by a contact prediction model that predicts contact between deformable and rigid bodies, with the aim of providing robots with the aforementioned human-like planning skill in order to efficiently automate downstream deformable object manipulation tasks such as cleaning dishes (Fig. \ref{fig:title}). Instead of \textit{contact reasoning}, in this paper we tackle the concept of \textit{contact prediction} of a 3D deformable tool acting on rigid objects, which is better suited for planning purposes. We take a data-driven approach with physics-based simulation to model the interactions between 3D deformable objects and rigid objects. We then use PointNet \cite{qi2016pointnet} architecture to form a mapping from point-cloud observation of the target object, and pose of the deformable tool to 3D representation of the contact points between the two bodies. The trained contact prediction model is then used as the driving force behind the planning of a subsequent task. Finally, we deploy SPONGE in the real world to demonstrate that the contact prediction model trained only with synthetic data from physics-based simulation can help to produce an efficient plan for a manipulation task to be successfully executed in the real world.

In summary, the main contributions of this paper are as follows:
\begin{itemize}
    \item A deep learning-based contact prediction model that predicts the contacts between 3D deformable and rigid objects under interactions.
    \item A planning pipeline powered by the proposed contact prediction model to efficiently automate deformable object manipulation tasks.    
    \item A novel experimental dataset containing 3D contact locations, 3D net force vectors, ands labeled contact areas when pressing deformable objects against rigid objects. The dataset was collected from 10 rigid objects interacting with a deformable sponge, with more than ten thousand data points.    
    \item A thorough empirical evaluation of the proposed method, both in simulation and on real hardware, demonstrating the ability of the planning pipeline to generate high-quality trajectories that efficiently cover the entire desired area to accomplish the manipulation task.
\end{itemize}

\section{Related work}
\label{sec:related_work}
To put our work in context, we next review three complementary viewpoints, the interaction between deformable and rigid objects, planning for deformable objects, and the simulation of deformable objects.
\subsection{Interaction between deformable and rigid objects}
Recently, the robotic research community has made great strides in learning complex dynamics of deformable objects, which in turn enable robots to achieve impressive results in sophisticated manipulation tasks such as folding \cite{hietala2022closing,speedfolding}/ unfolding clothes \cite{lin2021learning, ha2021flingbot}, untangling rope \cite{Grannen2020UntanglingDK} and manipulating deformable objects in target configurations \cite{shen2022acid}. Nevertheless, instead of only looking at the behavior of deformable objects, various methods have been proposed to study the interaction between rigid and deformable objects and take advantage of this knowledge to facilitate different manipulation tasks in different sectors such as grasp generation \cite{Brahmbhatt_2020_ECCV,jiang2021graspTTA,defggcnn}, assistive dressing \cite{erickson17,erickson18,wang2022visual}, assistive surgery \cite{Haouchine18}, or food processing \cite{heiden2021disect,sundaresan2022learning,lin2022planning}. 


Previous works attempted to estimate the interaction between deformable and rigid objects using real-world data \cite{Haouchine18,sundaresan2022learning,jiang2021graspTTA,Brahmbhatt_2020_ECCV}, while recent works sought to study this interaction using synthetic data from simulation. For example, Erickson \etal \cite{erickson17} presented a data-driven method trained in physics-based simulation that can infer the force applied onto a person's body from end-effector measurements. The inferred force is used along with model predictive control (MPC) to improve assistive dressing \cite{erickson18}. This idea has recently been leveraged for better generalization across objects and assistive tasks in \cite{wang2022visual}. Specifically, instead of using end effector measurements, Wang \etal \cite{wang2022visual} proposed a visual haptic reasoning method that uses point-cloud observations to estimate the contact force distribution when manipulating cloth on rigid objects. The model was experimentally proven to be reasonably transferred from simulation to different assistive tasks in the real world. It is worth noting that our work is highly inspired by \cite{wang2022visual}, but instead of doing \textit{contact reasoning} for cloth, our work in this paper aims at tackling \textit{contact prediction} for volumetric deformable objects, where our goal is to attempt to predict the contact areas in advance. 
\subsection{Planning for Deformable Object Manipulation}
Planning manipulation tasks involving the interaction of deformable and rigid objects is, compared to that involves only deformable objects, a less explored research area. Only until recently have a few works proposed methods to plan tasks either for rigid tools and deformable objects such as food processing \cite{lin2022planning,Seita2022toolflownet}, or for cloth and rigid objects such as assistive dressing \cite{wang2022visual,erickson18}. Continuing in this line of research, instead of focusing on cloth, in this paper we aim to learn the interaction between volumetric deformable tools and rigid objects and exploit this knowledge in planning manipulation tasks such as cleaning tasks.

\subsection{Deformable Object Simulation}
Research in computer graphics has made significant progress in the development of simulations of deformable objects \cite{Andrew06}. This progress has, in turn, driven recent simulators to manipulate deformable objects. Over the past few years, different physics-based simulators have been developed that allow us to simulate the interaction between deformable and rigid objects for various types of deformable objects, ranging from cloth \cite{clothsim_19, softgym,Qiao2020Scalable}, liquid \cite{softgym,Schenck18}, 3D deformable objects \cite{heiden2021disect,isaacgym}. For example, Wang \etal \cite{wang2022visual} trained the aforementioned visual haptic reasoning model using the data generated by the SoftGym simulator, and the result showed that the model is also well transferred to the real world.


In this work, since our goal is to study the interaction between volumetric deformable objects and rigid objects, we use the existing Isaac Gym simulator to develop our desired simulation environment. The main reason behind this choice is that Isaac Gym has been shown to provide accurate simulation of volumetric deformable objects, which, in turn, powered different downstream robotic tasks such as grasp analysis \cite{defgraspsim,Kim2021IPCGraspSimRT,metrics2022}, grasp synthesis \cite{defggcnn} or learning policies for robotic assembly \cite{Narang2022FactoryFC}.

\section{Problem Formulation}
\label{sec:prob_form}

\begin{figure}[!t]
    \centering
    \def\svgwidth{.6\linewidth}
     {\fontsize{10}{10}
\begingroup%
  \makeatletter%
  \providecommand\color[2][]{%
    \errmessage{(Inkscape) Color is used for the text in Inkscape, but the package 'color.sty' is not loaded}%
    \renewcommand\color[2][]{}%
  }%
  \providecommand\transparent[1]{%
    \errmessage{(Inkscape) Transparency is used (non-zero) for the text in Inkscape, but the package 'transparent.sty' is not loaded}%
    \renewcommand\transparent[1]{}%
  }%
  \providecommand\rotatebox[2]{#2}%
  \newcommand*\fsize{\dimexpr\f@size pt\relax}%
  \newcommand*\lineheight[1]{\fontsize{\fsize}{#1\fsize}\selectfont}%
  \ifx\svgwidth\undefined%
    \setlength{\unitlength}{360bp}%
    \ifx\svgscale\undefined%
      \relax%
    \else%
      \setlength{\unitlength}{\unitlength * \real{\svgscale}}%
    \fi%
  \else%
    \setlength{\unitlength}{\svgwidth}%
  \fi%
  \global\let\svgwidth\undefined%
  \global\let\svgscale\undefined%
  \makeatother%
  \begin{picture}(1,1.01320941)%
    \lineheight{1}%
    \setlength\tabcolsep{0pt}%
    \put(0,0){\includegraphics[width=\unitlength,page=1]{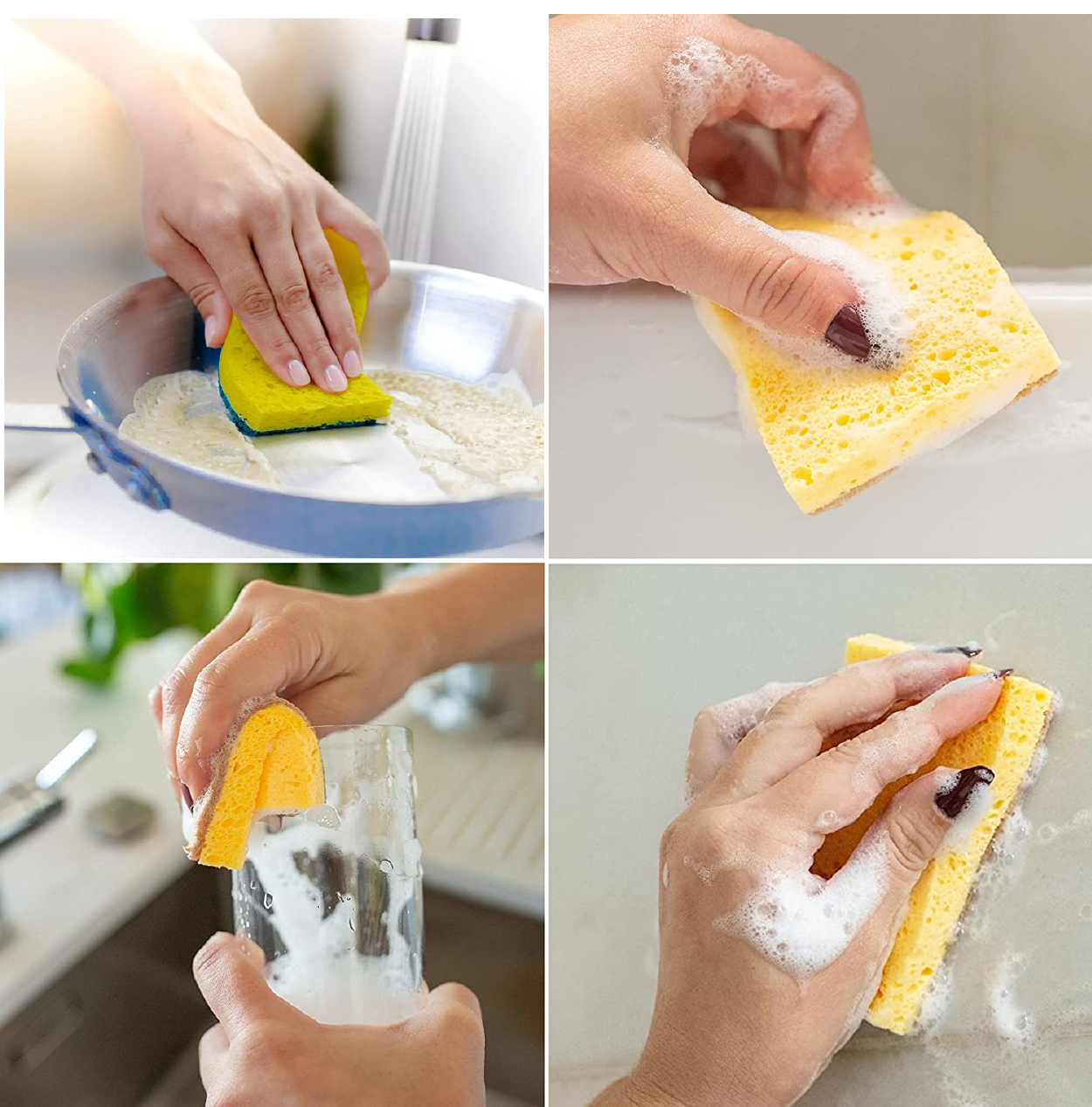}}%
    \put(0,0.95){\color[rgb]{0,0,0}\makebox(0,0)[lt]{\lineheight{1.25}\smash{\begin{tabular}[t]{l}a)\end{tabular}}}}%
    \put(0.5,0.95){\color[rgb]{0,0,0}\makebox(0,0)[lt]{\lineheight{1.25}\smash{\begin{tabular}[t]{l}b)\end{tabular}}}}%
    \put(0,0.45){\color[rgb]{0,0,0}\makebox(0,0)[lt]{\lineheight{1.25}\smash{\begin{tabular}[t]{l}c)\end{tabular}}}}%
    \put(0.5,0.45){\color[rgb]{0,0,0}\makebox(0,0)[lt]{\lineheight{1.25}\smash{\begin{tabular}[t]{l}d)\end{tabular}}}}%
  \end{picture}%
\endgroup%
} 
    \caption{Different tactics of human to create contact with various complex surface profiles using a deformable sponge.}
    \label{fig:idea}
    \vspace{-0.5cm}
\end{figure}
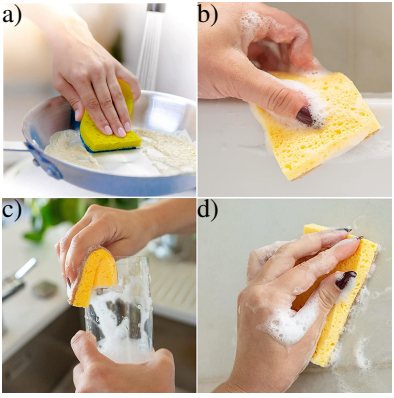

In this work, we focus on manipulation tasks that involve the interaction between a volumetric deformable tool and rigid objects, such as the one shown in \figref{fig:idea}. Specifically, this work addresses the problem of area coverage planning under deformations. The goal is to generate an optimal trajectory to cover the entire surface of a rigid object with a deformable tool while taking into account the interaction of the two bodies. To achieve this, we first need to learn the correlation between the geometric features of rigid objects and the contact areas between a deformable tool and those rigid objects. 
More formally, we train a model $\mathcal{M}$ that takes as input a point cloud of the target object $\mathcal{P}_O$ and the contact point $p_c$ ($p_c \in \mathcal{P}_O$) , and produces a contact map $\mathcal{P}_C$ for $N$ object points $(p_i \in \mathcal{P}_O)_{i=1}^N$.
\begin{equation*}
    \mathcal{M}: (\mathcal{P}_O,p_c) \mapsto \mathcal{P}_C.
\end{equation*} 
The learned model $\mathcal{M}$ is then used as input for an area coverage trajectory planning algorithm to generate an optimal trajectory consisting of $K$ contact points in an optimized ordered $\mathcal{T}_{opt} = \{p_{c_i}\}_{i=1}^K$ that achieves full area coverage of the target objects with the deformable tool in the most efficient way.
\section{Method}
\label{sec:method}
\begin{figure*}[!t]
    \centering
    \def\svgwidth{.9\linewidth}
     {\fontsize{10}{10}
    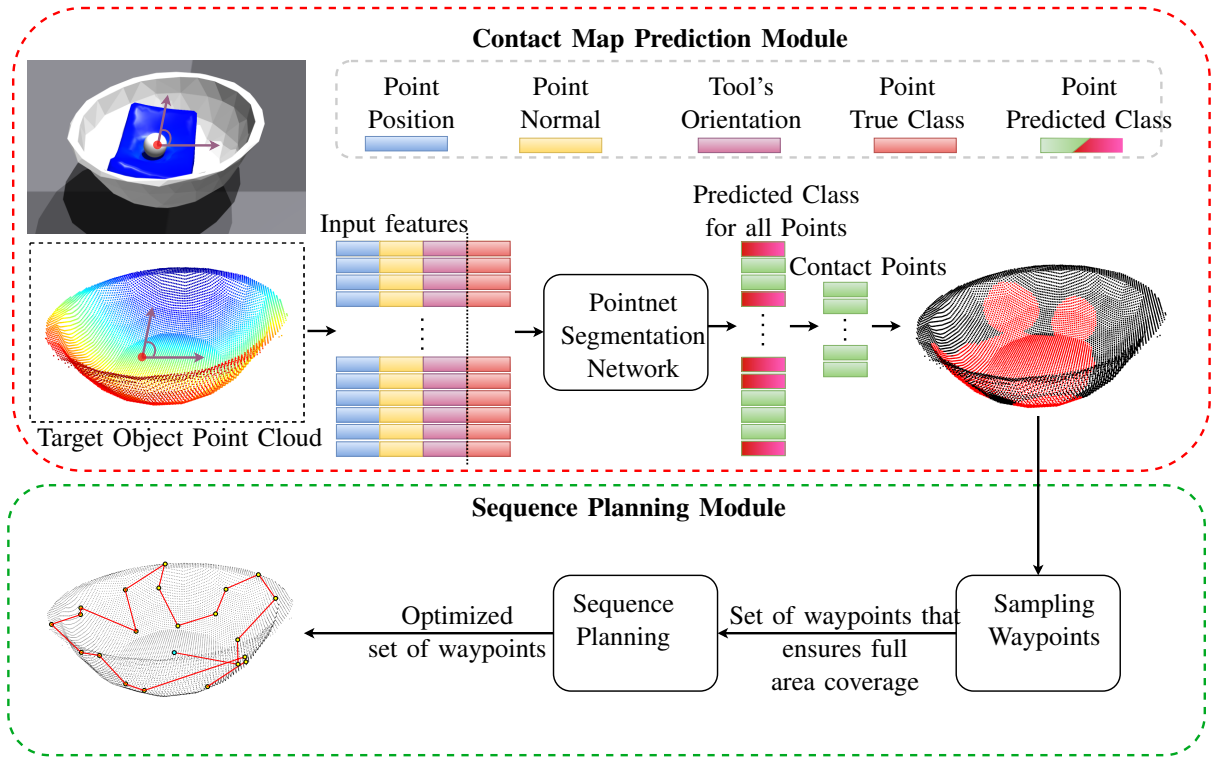} 
    \caption{The proposed planning pipeline consists of: a contact map prediction module learns from target object point clouds, which form the input to a dense point cloud network to produce per-point contacts; a sequence planning module that harnesses the trained prediction model to generate an optimal trajectory to accomplish the task.}
    \label{fig:pipeline}
    \vspace{-0.5cm}
\end{figure*}

The proposed planning pipeline shown in \figref{fig:pipeline} consists of two important steps:
\begin{enumerate*}[label=(\roman*)]
    \item learning to predict the contact between the deformable tool and target objects,
    \item planning an area-coverage trajectory on the target objects.
\end{enumerate*}
\subsection{Contact Map Prediction Model}
Knowledge of the contact area between two bodies is crucial when it comes to planning manipulation tasks associated with the interaction between two bodies. Let us consider \figref{fig:idea}, in which a human manipulates a deformable sponge to clean rigid dishes. From the figure, we can see that the contact areas between the sponge and the dishes are highly dependent on the contact location and the geometric features at that contact location. For example, in the case of objects with curved surfaces (\figref{fig:idea}~a,b,c), the deformability of the sponge allows it to conform to the curved surface to cover more area of the object. Let us examine \figref{fig:idea}~a, one can wipe both the bottom and the wall of the pan simply by pressing the sponge on the intersection line, which would not be possible if the sponge is rigid. Inspired by this behavior, we want to develop a model that learns a mapping from the geometric features of the rigid objects to the contact areas between the deformable and the rigid objects.

To this end, we propose using a dense point cloud network to model the contact information between the deformable tool and the target object. Specifically, we use a Pointnet segmentation network \cite{qi2016pointnet}, which, given an input point cloud of the target object $\mathcal{P}_O$ produces per-point outputs. It is worth noting that we are not training the Pointnet segmentation network to do segmentation, as the name implies, but to predict per-point contact class, which indicates whether a point of the target object is in contact with the deformable tool or not. 

The input of the network includes the position and normal vector of each point $p_i \in \mathcal{P}_O$ and a feature vector associated with each point $p_i$. Point positions are normalized to the zero mean, enabling the model to be invariant for point-cloud translations. The feature vector is a two-dimensional vector [$sin(\theta), cos(\theta)$] representing the orientation of the deformable tool around the Z axis at the contact location $p_c$. For points that are not the contact location, the feature vectors are defined as [0,0]. The output of the contact prediction model is the contact class of each point $p_i$ of the target object, where 1 is in contact and 0 is not in contact with the deformable tool. The proposed network is trained with supervised learning manner on a synthetic dataset with the Binary Classification Loss funtion. The synthetic training dataset is explained further in \secref{sec:dataset}.

\subsection{Area Coverage Planning Under Deformations}
As stated in \secref{sec:prob_form}, in this paper we address the problem of area coverage planning under deformations, where our goal is to plan an optimal trajectory that covers the entire surface of the target objects using a deformable tool while harnessing the learned contact map prediction model. The proposed algorithm consists of two steps: 
\begin{enumerate*}
    \item sampling waypoints, where we solve the Set Cover problem to sample sets of waypoints that ensure 100\% of the deformable tool's area coverage of the target objects, 
    \item sequence planning, where we choose and optimize the optimal trajectory from the obtained sets of waypoints. 
\end{enumerate*}
A summary of the algorithm for the sampling waypoints step is shown in Algorithm \ref{alg:planning}.

{\centering
\begin{minipage}{\linewidth}
 \removelatexerror
  \begin{algorithm}[H]
    \caption{Sampling sets of waypoints}\label{alg:planning}
    \SetKwInOut{Input}{Input}   
    \SetKwInOut{Output}{Output}
    \Input{Target object point cloud $\mathcal{P}_O$, number of sets to be sampled $n_{sets}$} 
    \Output{$n_{sets}$ sets of waypoints $\mathcal{T} = \{[p_{i_1} \dots p_{i_n}],[p_{i_1} \dots p_{i_n}],\dots\}$}
    \For {$j = 0, \dots n_{sets}$}{
        \begin{enumerate}
            \item Sample a random contact point $p_{i} \in \mathcal{P}_O$ and \\ gripper orientation $\theta_i$
            \item Predict contact map at $p_{i}: \mathcal{P}_{Ci} = \mathcal{M}(\mathcal{P}_O,p_{i},\theta_i)$
            \item Append contact point $p_i$ to $\mathcal{T}_j$
            \item Remove all points that are in contact with the \\  original target point cloud: $\tilde{\mathcal{P}_O} = \mathcal{P}_O - \mathcal{P}_{Ci}$
        \end{enumerate}
            \Repeat{$\tilde{\mathcal{P}_O} =\emptyset$}{%
            \text{Steps 1 to 3.} 
            \text{Update $\tilde{\mathcal{P}_O} = \tilde{\mathcal{P}_O} - \mathcal{P}_{Ci}$}
                    }
        }
  \end{algorithm}
\end{minipage}
\par
}
More specifically, the planning algorithm takes the point cloud of the target object along with the number of sets to be sampled and produces $\mathcal{T}$ containing $n_{sets}$  sets of contact points on the surface of the target object. We achieve this by solving the set cover problem using a heuristic bottom-up sampling algorithm, where we first randomly sample a contact point on the target object surface, predict the contact areas at that point, and remove all the points that are in contact from the target object point cloud. This process is repeated until the remaining point cloud of the target object is empty, indicating that we have covered the entire object. 

Once $\mathcal{T}$ is obtained, we proceed to the sequence planning step, where the goal is to produce an optimal trajectory that achieves full area coverage of the target objects while minimizing a certain cost measure, such as the travel distance. We frame this problem in relation to the well-known travel salesman problem (TSP). We achieve this by solving the TSP with the 2-Opt algorithm for the best set of waypoints $\mathcal{T}_{j} \in \mathcal{T}$, which ensures 100\% of the area coverage of the target objects with the least number of waypoints. More formally, the optimal trajectory is defined as
\begin{equation}
    \mathcal{T}_{opt} =  TSP(\mathop{\arg \min}\limits_{\mathcal{T}_j \in \mathcal{T}} len(\mathcal{T}_j))
\end{equation}

\section{Dataset}
\label{sec:dataset}
To train the contact map prediction model, we need a dataset consisting of point cloud of the rigid object, and contact areas when they undergo interaction with the volumetric deformable object. To date, there exists no such dataset, and thus we opted to curate our own synthetic dataset. To achieve this, we develop a simulation environment, named SPONGESim. The simulation environment and the dataset generation process are explained further below.

\begin{figure}[!t]
    \centering
    \def\svgwidth{\linewidth}
     {\fontsize{10}{10}
\begingroup%
  \makeatletter%
  \providecommand\color[2][]{%
    \errmessage{(Inkscape) Color is used for the text in Inkscape, but the package 'color.sty' is not loaded}%
    \renewcommand\color[2][]{}%
  }%
  \providecommand\transparent[1]{%
    \errmessage{(Inkscape) Transparency is used (non-zero) for the text in Inkscape, but the package 'transparent.sty' is not loaded}%
    \renewcommand\transparent[1]{}%
  }%
  \providecommand\rotatebox[2]{#2}%
  \newcommand*\fsize{\dimexpr\f@size pt\relax}%
  \newcommand*\lineheight[1]{\fontsize{\fsize}{#1\fsize}\selectfont}%
  \ifx\svgwidth\undefined%
    \setlength{\unitlength}{1579.5bp}%
    \ifx\svgscale\undefined%
      \relax%
    \else%
      \setlength{\unitlength}{\unitlength * \real{\svgscale}}%
    \fi%
  \else%
    \setlength{\unitlength}{\svgwidth}%
  \fi%
  \global\let\svgwidth\undefined%
  \global\let\svgscale\undefined%
  \makeatother%
  \begin{picture}(1,0.46438746)%
    \lineheight{1}%
    \setlength\tabcolsep{0pt}%
    \put(0,0){\includegraphics[width=\unitlength,page=1]{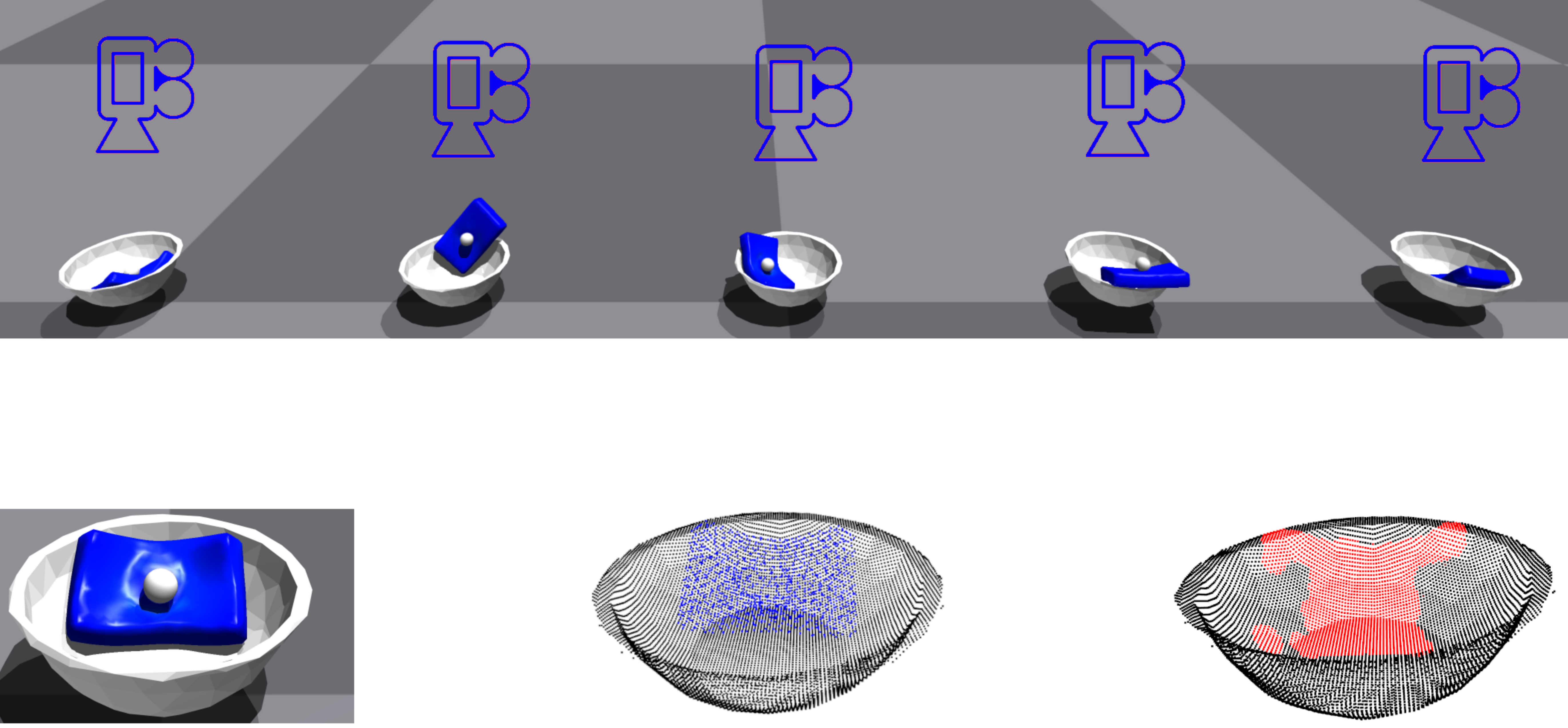}}%
    \put(0.75,0.2){\makebox(0,0)[lt]{\lineheight{1.25}\smash{\begin{tabular}[t]{c}Ground-truth \\contact map\end{tabular}}}}%
    \put(0.3,0.2){\makebox(0,0)[lt]{\lineheight{1.25}\smash{\begin{tabular}[t]{c}Deformable tool's nodal \\postion \& contact forces\end{tabular}}}}%
    \put(0,0){\includegraphics[width=\unitlength,page=2]{sponge_env.pdf}}%
    \put(0.02,0.2){\makebox(0,0)[lt]{\lineheight{1.25}\smash{\begin{tabular}[t]{c}End of \\interaction\end{tabular}}}}%
  \end{picture}%
\endgroup%
} 
    \caption{We introduce SPONGESim, a manipulation environment of realistic volumetric deformable objects interacting with rigid objects. Top: example images from the five parallel environments performing five different interactions at randomized contact locations. Bottom: the process from simulating the interaction between two objects to getting ground-truth contact map for the network.}
    \label{fig:sponge_env}
    \vspace{-0.3 cm}
\end{figure}

\textbf{\textit{SPONGESim} environment:} The developed SPONGESim contains actions and sensor simulation data for interacting between volumetric deformable objects and rigid objects. More precisely, we use a physics-based simulator, specifically the NVIDIA Isaac Gym simulator with Flex engine, which features the GPU-accelerated finite element method (FEM) which represents a volumetric deformable body as a graph of connected tetrahedrons. As shown in \figref{fig:sponge_env}, we build an environment in which the robot uses a deformable tool attached to a hemispherical finger to interact with the surface of target objects. In this environment, we first randomly sample contact points on the surface of the target objects and vary the orientation of the tool around the Z axis. The deformable tool then approaches the contact points from their normal directions and applies 5N of force to the target objects. Once the desired contact force is reached, we store the point cloud, contact points, deformable tool orientation, and ground-truth contact map between the deformable tool and the object as data points for training. This process of getting this information is further discussed in the following. 

\textbf{Inputs to the network:} We captured depth images of target objects with a virtual camera set to view the scene from the top down. The depth images are then transformed into point clouds of the target objects. The point clouds of the target objects along with the sampled contact points and deformable tool orientation are then stored to serve as input of the contact map prediction network.

\textbf{Ground-truth contact map:} During the interaction between rigid and deformable bodies, Isaac Gym only returns the nodal positions, the nodal contact forces of the deformable body. Thus, the contact map between the rigid and deformable objects is not available to be stored directly. Therefore, we need to calculate the desired contact map from the obtained nodal position and nodal contact forces of the deformable tool. To do this, we first set a threshold of 0.5N on the nodal contact forces of the deformable tool so that a point is only considered to be in contact with the target objects if its contact force is greater than the threshold. This gives us a set of points of the deformable tool that are in contact with the target object. Next, for each point of the deformable tool in the contact set, we find its closest points (within a distance $\tau$) on the surface of the target object and label those points as in contact with the deformable tool. The labeled point cloud is then used as the ground-truth contact map for training the contact prediction network. 

\begin{figure}[!t]
    \centering
    \def\svgwidth{\linewidth}
     {\fontsize{10}{10}
    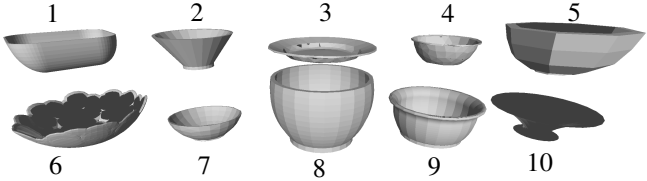} 
    \caption{The 10 individually numbered rigid objects used in simulation. The objects represent a high variation in size, shape, and curvatures.}
    \label{fig:sim_objs}
    \vspace{-0.7cm}
\end{figure}

\textbf{Training dataset:} As a training dataset, we generate and label the contact maps for 10 objects shown in \figref{fig:sim_objs}. The objects include 2 objects from the YCB dataset \cite{ycb}, and 8 objects from the ShapeNet dataset \cite{shapenet}. The objects chosen are mainly bowls and plates with different curvatures. For each object, labeled data was collected by simulating 1000 contacts between the deformable tool and the object. We further sped up the data generation and labeling process by executing contact interactions in parallel environments. In total, tens of thousands of unique contact interactions were conducted. 70\% of contact interactions were assigned for training, 15\% for validation, and 15\% for testing.


\section{{Experiments and Results}}
\label{sec:exp_and_res}
We evaluate each component of our \textit{SPONGE} pipeline: contact map prediction and sequence planning. In particular, our aim is to answer the following key questions:
\begin{itemize}
    \item Can the proposed contact prediction model predict the contact map between the deformable tool and the target objects given only the pose of deformable tool?
    \item Is the optimal trajectory generated by \textit{SPONGE} capable of covering the entire surface of the target objects in the simulation? 
    \item Can \textit{SPONGE}, trained purely on synthetic data, generate a good trajectory to accomplish the task in the real world?
\end{itemize}
\subsection{Contact Map Prediction Result}
\begin{figure}[!t]
    \centering
    \def\svgwidth{\linewidth}
     {\fontsize{10}{10}
\begingroup%
  \makeatletter%
  \providecommand\color[2][]{%
    \errmessage{(Inkscape) Color is used for the text in Inkscape, but the package 'color.sty' is not loaded}%
    \renewcommand\color[2][]{}%
  }%
  \providecommand\transparent[1]{%
    \errmessage{(Inkscape) Transparency is used (non-zero) for the text in Inkscape, but the package 'transparent.sty' is not loaded}%
    \renewcommand\transparent[1]{}%
  }%
  \providecommand\rotatebox[2]{#2}%
  \newcommand*\fsize{\dimexpr\f@size pt\relax}%
  \newcommand*\lineheight[1]{\fontsize{\fsize}{#1\fsize}\selectfont}%
  \ifx\svgwidth\undefined%
    \setlength{\unitlength}{3369.37915231bp}%
    \ifx\svgscale\undefined%
      \relax%
    \else%
      \setlength{\unitlength}{\unitlength * \real{\svgscale}}%
    \fi%
  \else%
    \setlength{\unitlength}{\svgwidth}%
  \fi%
  \global\let\svgwidth\undefined%
  \global\let\svgscale\undefined%
  \makeatother%
  \begin{picture}(1,0.63134459)%
    \lineheight{1}%
    \setlength\tabcolsep{0pt}%
    \put(0,0){\includegraphics[width=\unitlength,page=1]{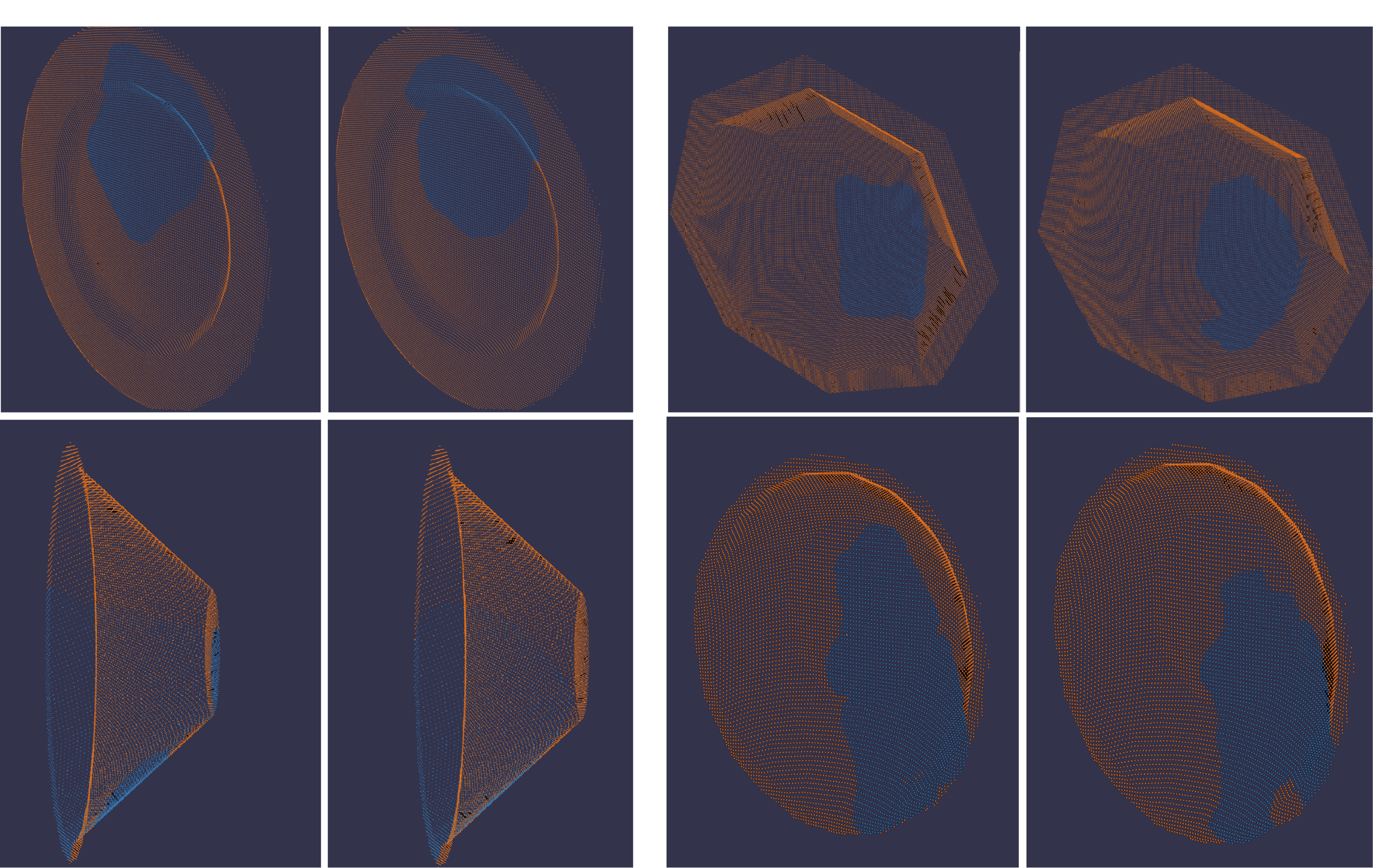}}%
    \put(0.02,0.61790652){\makebox(0,0)[lt]{\lineheight{1.25}\smash{\begin{tabular}[t]{l}Ground-truth\end{tabular}}}}%
    \put(0.28,0.61790652){\makebox(0,0)[lt]{\lineheight{1.25}\smash{\begin{tabular}[t]{l}Prediction\end{tabular}}}}%
    \put(0.52,0.61790652){\makebox(0,0)[lt]{\lineheight{1.25}\smash{\begin{tabular}[t]{l}Ground- truth\end{tabular}}}}%
    \put(0.8,0.61790652){\makebox(0,0)[lt]{\lineheight{1.25}\smash{\begin{tabular}[t]{l}Prediction\end{tabular}}}}%
  \end{picture}%
\endgroup%
} 
    \caption{Visualization of the contact map predictions on the test dataset and the ground-truth in simulation. Left columns show the ground-truth and the right columns show the prediction. Blue indicates points that are in contact with the deformable tool, while orange denotes points that are not in contact with the deformable tool. \bestcolor}
    \label{fig:pred_result}
    \vspace{-0.5cm}
\end{figure}
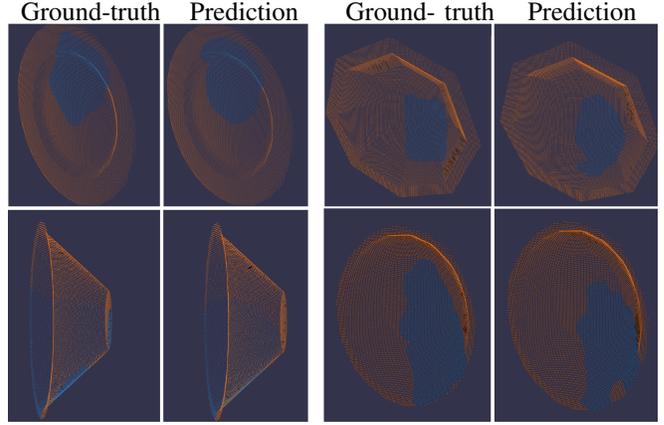

We access the performance of the proposed prediction model based on the contact prediction F1 score on the test dataset, which is approximately \textbf{0.95}. This high score indicates that the proposed model is capable of accurately predicting the contact map, given only the applied position and the rotation of the sponge. \figref{fig:pred_result} qualitatively compares the predicted contact map from the proposed model with the ground truth in the simulation. As shown, the predicted contact maps are qualitatively similar to ground truth. It should be noted that the model was able to capture the correlation between the geometric features of the target object and the contact map between the two bodies. For example, let us examine the bowls shown in \figref{fig:pred_result}, because of the deformability of the tool, it conforms to the curvature of the bowls when applied to the side of the bowl. Our proposed model was able to capture this behavior by taking into account the pose of deformable tool, and the local features close to the applied location.

\begin{figure}[!t]
    \centering
    \includegraphics[width=.6\linewidth]{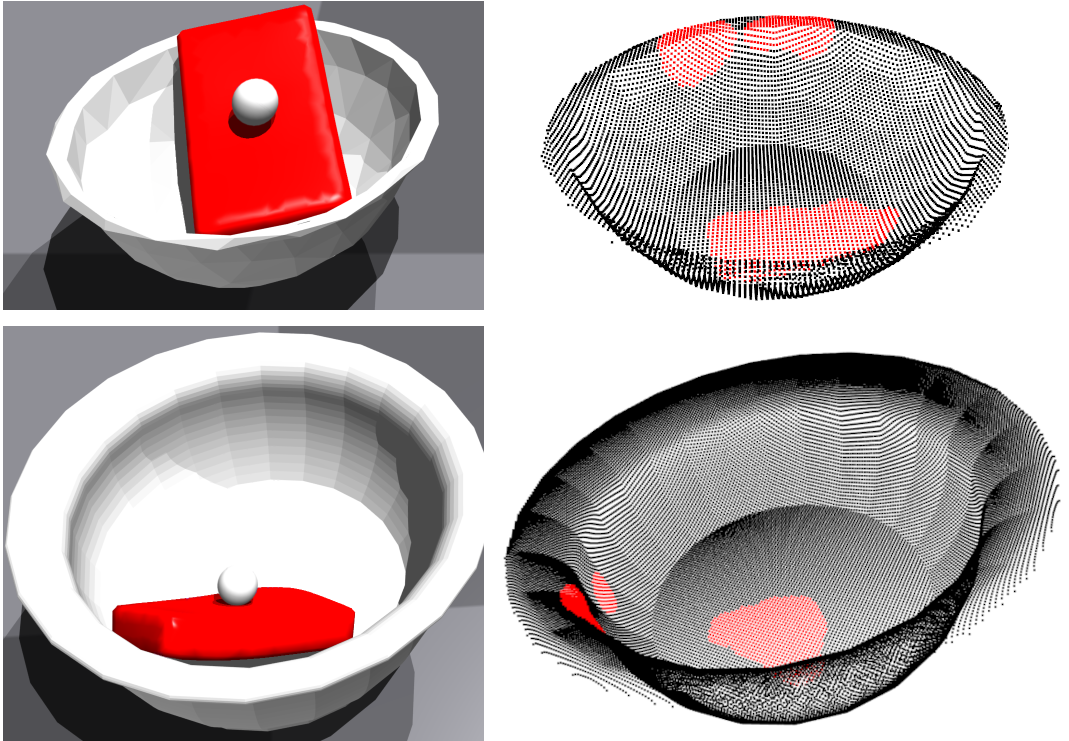}
    \caption{Visualization of labeled contact map when the sponge is set to be rigid.}
    \label{fig:rigid}
    \vspace{-0.5cm}
\end{figure}

It is worth mentioning that to single out the effect of deformability has on the planning pipeline, we attempted to learn a contact map prediction model in the case that the tool is rigid. To do that, we generated a new dataset following the same procedure mentioned in \secref{sec:dataset}, but in this case, we assume the tool is rigid by setting its Young’s modulus to $2\cdot10^{9}$. However, the model was unable to learn the correlation from the geometric features of the target objects and the contact map in this case due to the sparse ground truth, as shown in \figref{fig:rigid}. Since the tool is rigid, it does not conform to the surface of the target objects, which in turn leads to a sparsity in the ground truth. 
\subsection{Planning in Simulation Result}
We investigate the quality of the generated trajectories in the context of the dish cleaning task in the Isaac Gym simulator using the same environment and the same ten target objects shown in \figref{fig:sim_objs}. Due to the limited options regarding the controller for soft bodies in Isaac Gym, we simply execute the planned trajectory using a Cartesian position controller. The position and contact forces of the sponge are recorded during the execution process. Then, based on the recorded data, contact maps between the target objects and the sponge are calculated using the same approach described in \secref{sec:dataset}. As the objective of the task is to cover the entire surface of the target objects with the planned trajectory, we quantify the quality of the trajectory by the area coverage (\textit{i.e.,~}proportion of the contact points to the total population of point clouds). The higher and closer to 100 the proportion, the better the quality of the planned trajectory. For each object, we randomly initialized its starting position 20 times and evaluated the best trajectory among 50 sampled ones. In total, we evaluated 200 trajectories on all ten objects.
\begin{table}
    \centering
	\ra{1}\tbs{10}
	\caption{\label{tb:simexp} The average area coverage (\%) and average number of waypoints over 20 trajectories on the target objects in simulation. $\uparrow$: higher the better}
    \begin{tabular}{@{}lcc@{}}
        \toprule
        Object & \multicolumn{1}{l}{Area Coverage} & \multicolumn{1}{l}{Number of Waypoints} \\
        \midrule
        1  & 89.5    &  19.3  \\
        2  & 97.3    &   13.3\\
        3  &  100   & 18.1  \\
        4  &  96.2    & 14.1  \\
        5 &  87.6    &  23.6\\
        6  & 97.9    &   22.3\\
        7  & 98.7    &   12.4\\
        8  & 88.5    &  32.7 \\
        9  & 92.1    &   22.4\\
        10 &  94.6    &  20.2\\
        \midrule
        All  $\uparrow$ & 94.27    & \_  \\
        \bottomrule
    \end{tabular}
     \figvspace{}
\end{table}

\tabref{tb:simexp} presents the average area coverage over 20 optimal trajectories on all objects. These results clearly show that the proposed planning pipeline is capable of producing high-quality trajectories that cover approximately 94\% the surfaces of different objects with varying geometry and curvatures. Although there exist some declines in the area coverage of objects 1, 5 and 8 compared to other objects, the coverage is still above 87\% for all three objects. Based on our observations, one reason for that is due to the fact that we are using only a position controller to execute the trajectories; sometimes, for these large-sized objects with large curvatures, the controller does not necessarily apply enough contact force at certain locations, which in turn leads to the incorrect contact point calculation.

Furthermore, the average number of waypoints for 20 trajectories over all objects is also reported in \tabref{tb:simexp}. As expected, the number of waypoints needed to cover the entire surface increases as the size of 
the target objects increases. It is exciting that for objects 2 and 4, it takes only less than 15 waypoints to accomplish the task, which shows the effectiveness of the proposed planning pipeline.
\subsection{Real Robot Deployment}
\begin{figure*}[!t]
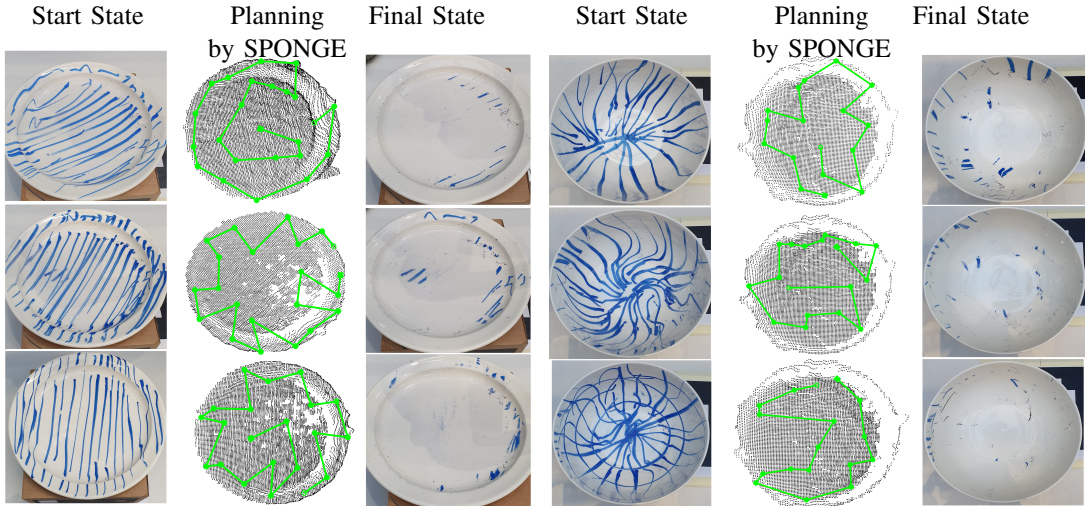

    \centering
    \begin{tabular}{c@{}c@{}}
        \def\svgwidth{.4\linewidth}
         {\fontsize{10}{10}
\begingroup%
  \makeatletter%
  \providecommand\color[2][]{%
    \errmessage{(Inkscape) Color is used for the text in Inkscape, but the package 'color.sty' is not loaded}%
    \renewcommand\color[2][]{}%
  }%
  \providecommand\transparent[1]{%
    \errmessage{(Inkscape) Transparency is used (non-zero) for the text in Inkscape, but the package 'transparent.sty' is not loaded}%
    \renewcommand\transparent[1]{}%
  }%
  \providecommand\rotatebox[2]{#2}%
  \newcommand*\fsize{\dimexpr\f@size pt\relax}%
  \newcommand*\lineheight[1]{\fontsize{\fsize}{#1\fsize}\selectfont}%
  \ifx\svgwidth\undefined%
    \setlength{\unitlength}{4650bp}%
    \ifx\svgscale\undefined%
      \relax%
    \else%
      \setlength{\unitlength}{\unitlength * \real{\svgscale}}%
    \fi%
  \else%
    \setlength{\unitlength}{\svgwidth}%
  \fi%
  \global\let\svgwidth\undefined%
  \global\let\svgscale\undefined%
  \makeatother%
  \begin{picture}(1,0.88774194)%
    \lineheight{1}%
    \setlength\tabcolsep{0pt}%
    \put(0,0){\includegraphics[width=\unitlength,page=1]{real_res.pdf}}%
    \put(0.68,0.9){\makebox(0,0)[lt]{\lineheight{1.25}\smash{\begin{tabular}[t]{l}Final State\end{tabular}}}}%
    \put(0,0){\includegraphics[width=\unitlength,page=2]{real_res.pdf}}%
    \put(0.05,0.9){\makebox(0,0)[lt]{\lineheight{1.25}\smash{\begin{tabular}[t]{l}Start State\end{tabular}}}}%
    \put(0.38,0.9){\makebox(0,0)[lt]{\lineheight{1.25}\smash{\begin{tabular}[t]{c}Planning \\ by SPONGE\end{tabular}}}}%
  \end{picture}%
\endgroup%
} 
	&	\def\svgwidth{.4\linewidth}
         {\fontsize{10}{10}
\begingroup%
  \makeatletter%
  \providecommand\color[2][]{%
    \errmessage{(Inkscape) Color is used for the text in Inkscape, but the package 'color.sty' is not loaded}%
    \renewcommand\color[2][]{}%
  }%
  \providecommand\transparent[1]{%
    \errmessage{(Inkscape) Transparency is used (non-zero) for the text in Inkscape, but the package 'transparent.sty' is not loaded}%
    \renewcommand\transparent[1]{}%
  }%
  \providecommand\rotatebox[2]{#2}%
  \newcommand*\fsize{\dimexpr\f@size pt\relax}%
  \newcommand*\lineheight[1]{\fontsize{\fsize}{#1\fsize}\selectfont}%
  \ifx\svgwidth\undefined%
    \setlength{\unitlength}{4530bp}%
    \ifx\svgscale\undefined%
      \relax%
    \else%
      \setlength{\unitlength}{\unitlength * \real{\svgscale}}%
    \fi%
  \else%
    \setlength{\unitlength}{\svgwidth}%
  \fi%
  \global\let\svgwidth\undefined%
  \global\let\svgscale\undefined%
  \makeatother%
  \begin{picture}(1,0.88015735)%
    \lineheight{1}%
    \setlength\tabcolsep{0pt}%
    \put(0,0){\includegraphics[width=\unitlength,page=1]{real_res_bowl.pdf}}%
	\put(0.68,0.9){\makebox(0,0)[lt]{\lineheight{1.25}\smash{\begin{tabular}[t]{l}Final State\end{tabular}}}}%
    \put(0.05,0.9){\makebox(0,0)[lt]{\lineheight{1.25}\smash{\begin{tabular}[t]{l}Start State\end{tabular}}}}%
    \put(0.38,0.9){\makebox(0,0)[lt]{\lineheight{1.25}\smash{\begin{tabular}[t]{c}Planning \\ by SPONGE\end{tabular}}}}%
  \end{picture}%
\endgroup%
} 
    \end{tabular}
    \caption{Qualitative visualizations of \textit{SPONGE} for real-world dish cleaning task. Columns headed by Start State are the target object with blue marker writings denoting dirt need to be removed. The optimal coverage trajectories planned by \textit{SPONGE} (solid green lines) are shown in columns headed by Planning by SPONGE. The columns headed by  Final State are when robot is done executing the early planned trajectories. \bestcolor}
    \label{fig:real_result}
    \vspace{-0.5cm}
\end{figure*}

To investigate how well the proposed pipeline performs in the real world, we conducted an experiment that performs a dish cleaning task with a \panda equipped with a hemispherical finger attached to a deformable sponge, as shown in \figref{fig:title}. The dimensions and material characteristics of the sponge used in the real-world experiment are identical to those of the sponge in the simulation. Two target objects used in this experiment are a bowl and a plate that are not from any dataset. This allows us to study whether the contact map predicted with the model trained only on synthetic data transfer to real objects and whether the trajectory generated by the planning pipeline is sufficient to accomplish the task in the real world. 

We used a Kinect v2 camera to capture the point cloud of the scene. We then segmented the object from the scene by subtracting
the background and the table from the obtained point cloud. The point cloud of the target object is then fed into the proposed planning pipeline to generate the optimal trajectory to clean the top surface of the object. Cartesian impedance control is employed to execute the generated trajectory to guarantee contact with the object surface. The robot first approaches each waypoint of the trajectory from its normal direction and then applies forces on the target object at that location for a short period of time before traversing to the next waypoint. It is important to note that the execution of the trajectory depends on the object that remains immobile during the process. In our case, the object is manually held to prevent the object from moving during execution. The problem could potentially be solved by using a dual-arm setup to remove the need for human intervention. 

As shown in \figref{fig:real_result}, the objective of the task is to clean all the blue marker writings, which represent dirt on the target objects. The performance of the proposed pipeline is then qualitatively and quantitatively evaluated based on the remaining amount of blue marker writings after the execution of the trajectories. Specifically, we quantitatively achieve this by first taking two RGB images of the target objects before and after the execution of the planned trajectories; then we calculate and compare the number of blue-color pixels between the two images. The coverage is then calculated as: 
\begin{equation*}
    \text{Coverage} = (1 - \frac{n_{after}}{n_{before}})*100
\end{equation*}
where $n_{before}, n_{after}$ denote the number of blue-color pixels of the target object before and after the execution of the trajectory, respectively. The higher and closer to 100 the coverage, the better the quality of the planned trajectory. For each object, we randomly placed it five times and evaluated the best coverage trajectory among 50 sampled ones, which in total amounts to 10 trajectories on two real-world objects.

\figref{fig:real_result} shows qualitative results of the real robot deployment of \textit{SPONGE} on the two target objects. From the results, we can see that even though there still remains some blue marker writings after the cleaning process, the robot has successfully accomplished the dish cleaning task by removing almost all of the blue marker writings from the surfaces of the target objects. This observation was further reflected by the quantitative results shown in \tabref{tb:realexp} with an impressive area coverage of more than 95\% in two different geometries. In addition, the plate requires 20 waypoints to cover its surface, whereas the bowl requires fewer waypoints due to the smaller size.   
\begin{table}
    \centering
	\ra{1}\tbs{10}
	\caption{\label{tb:realexp} The average area coverage (\%) and average number of waypoints over 5 trajectories on the target objects. $\uparrow$: higher the better}
    \begin{tabular}{@{}lcc@{}}
        \toprule
        Object & \multicolumn{1}{l}{Area Coverage} & \multicolumn{1}{l}{Number of Waypoints} \\
        \midrule
        Plate  & 94.5   &  20.4  \\
        Bowl  & 96.2    & 17.4  \\
        \midrule
        All  $\uparrow$ & 95.35    & \_  \\
        \bottomrule
    \end{tabular}
     \figvspace{}
\end{table}

\subsection{Discussion}
Despite the good results, several limitations of \textit{SPONGE} can be further investigated and improved in future work.

First, the current contact map prediction model lacks real-time knowledge of the en route contact map while moving from one contact point to another. For example, given two contact points A and B, the current model can only predict two separate contact maps associated with those two points. Therefore, the contact map created when moving the deformable object from point A to point B is unknown. One potential approach to this problem is to extend the current model by incorporating the robot kinematics, such as gripper velocity, into the model to learn the mapping from the robot movements to the contact map. Knowledge of the contact map when moving between contact points can potentially increase the efficiency of the planning pipeline, so that fewer waypoints would be needed to cover the entire surface of the objects.    

Second, in the sequence planning module, since we just randomly sample contact points until the entire surface is covered, the resulting trajectories may look counter-intuitive compared to that of humans. For example, humans usually clean dishes by executing a smooth spiral-shaped trajectory. Introducing a more systematic waypoints sampling approach may be the way to tackle this issue.

Last but not least, in this work, the planned trajectories are executed in an open-loop manner, where we omit the actual contact happening between the two objects during the manipulation. This information is important in reacting and adapting trajectories to uncertainties such as incorrect contact map prediction, or displacement of the target object during execution procedure. One interesting future work would be to introduce a contact reasoning model, which reasons the actual contact map based on the immediate observation, as a feedback to close the aforementioned control loop.

\section{Conclusions}
\label{sec:conclusions}

Studying the interaction between deformable and rigid objects is an interesting yet complicated problem due to the complexity in the modeling and simulating such contact. However, with the rapid development of physics-based simulators that support soft bodies, the research gap between rigid and deformable objects is becoming smaller. To leverage the capability of such simulators in this domain, we presented a planning pipeline for manipulation tasks involving the interaction between deformable and rigid objects, which we explicitly demonstrate in the dish cleaning task with a deformable sponge. The key component of the pipeline is the contact map prediction model trained entirely on synthetic data, which learns the mapping from the contact location to the contact map between the two objects. The output of the model is then used to generate the optimal trajectory that achieves full area coverage of the target objects. We demonstrate the performance of the planned trajectories through experiments in both simulation and real-world scenarios on objects with varying sizes and geometrical features. The results show that the proposed planning pipeline is capable of generating high-quality trajectories that efficiently achieve nearly full area coverage of the target objects with a deformable tool. Despite the good results, there is still room for improvements. Some limitations of the proposed pipeline that can be tackled are the lack of real-time knowledge of the en route contact map while traversing between two contact points, or counter-intuitive trajectories compared to those of humans. Another interesting future work would be to close the manipulation loop with real-time contact feedback to counteract uncertainties in the real world such as incorrect contact map prediction or displacement of objects while being manipulated.



\bibliographystyle{IEEEtran}
\bibliography{refs}

\end{document}